\documentclass[sigconf, nonacm]{acmart}

\AtBeginDocument{%
  }

\setcopyright{none}

\setcopyright{none}              
\renewcommand\footnotetextcopyrightpermission[1]{} 
\begin{document}

\title[ERIT: A Responsible Multimodal Affective Computing Benchmark for Older-Adult Emotion Recognition in the Wild]{ERIT: A Responsible Multimodal Affective Computing Benchmark for Older-Adult Emotion Recognition in the Wild}

\author{Rita Frieske}
\affiliation{
  \institution{The Hong Kong University of Science and Technology}
  \department{Department of Electronic and Computer Engineering}
  \city{Hong Kong}
  \country{Hong Kong}
}
\email{rmfrieske@connect.ust.hk}

\author{Bertram E. Shi} 
\affiliation{
  \institution{The Hong Kong University of Science and Technology}
  \department{Department of Electronic and Computer Engineering}
  \city{Hong Kong}
  \country{Hong Kong}
}
\email{eebert@ust.hk}

\renewcommand{\shortauthors}{Frieske and Shi}
\begin{abstract}
We introduce a controlled evaluation setting for image–text fusion in older adult emotion recognition, instantiated in the ERIT benchmark. Many existing video-based emotion datasets rely on frame sampling during evaluation, and differences in how frames are sampled can substantially affect results—yet such details are seldom disclosed, limiting reproducibility and fair comparison. ERIT removes this source of variability by pairing representative annotated frames with corresponding transcribed speech, so that each sample has fixed visual and textual inputs. The dataset covers seven emotions (anger, disgust, fear, happiness, sadness, surprise, and neutral), and is designed primarily as a controlled evaluation resource for the underrepresented population of older adults.
We evaluate ERIT in zero-shot settings with LLaVA-NeXT, Qwen2-VL, GPT-4V, and GPT-4o, and also fine-tune LLaVA-NeXT with LoRA, comparing results using strict, multi-label, and balanced accuracies. We show that multimodal fusion improves ambiguity-aware metrics but remains visually dominated under strict evaluation, exposing limitations of current VLMs for older-adult affect recognition. Robustness and label agreement experiments show how age-related facial morphology introduces misclassifications, and how cross-modal label disagreement (e.g., happy text / sad image) influences fusion results.

\end{abstract}





\maketitle

\section{Introduction}

Reliable multimodal emotion recognition is critical for human--computer interaction, where models must interpret affect from facial, linguistic, and other signals~\cite{Leong2023, Huang2023, Acheampong2020}. Recent work highlights the importance of understanding how different modalities contribute to model predictions~\cite{Zhao2023, Li2024, George2024}, particularly when signals are ambiguous or conflicting. Controlled evaluation is therefore essential for assessing whether systems genuinely integrate multimodal cues or rely on a single dominant modality.
\begin{figure}[t]
  \centering
  \includegraphics[width=\linewidth]{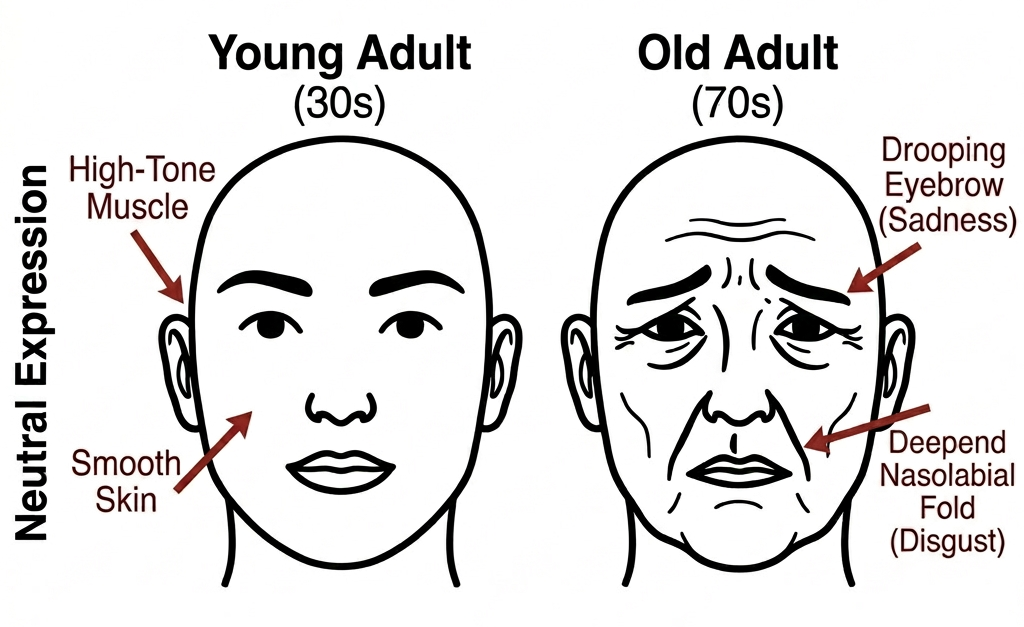}
  \caption{Age-related facial morphology may be misread by emotion-recognition models as active emotional expression, potentially biasing zero-shot predictions.}
  \Description{A side-by-side schematic comparison of a young adult (30s) and an old adult (70s) in a neutral expression. The young adult face is labeled as having high-tone muscle and smooth skin, resulting in a high-signal neutral baseline. The old adult face features red arrows pointing to a drooping eyebrow labeled as a sadness mimic and a deepened nasolabial fold labeled as a disgust mimic, resulting in a noisy, low-confidence neutral signal.}
  \label{fig:aging-morphology}
\end{figure}
Emotion recognition systems in human-centered settings must operate across diverse populations. Older adults represent an underrepresented group in existing emotion recognition datasets, despite their growing importance in applications such as healthcare and assistive technologies~\cite{Sharma2021, cahyawijaya23_interspeech}. Emotional expression in this demographic may differ from that of younger populations, introducing additional challenges (see Figure~\ref{fig:aging-morphology})~\cite{Ebner2009}.
Despite the availability of multimodal datasets, evaluation is often performed on video-based data where frame sampling is required. Several widely used benchmarks (e.g., CMU-MOSEI, IEMOCAP, CMU-MOSI) rely on uncontrolled or under-specified sampling strategies during evaluation ~\cite{bagher-zadeh-etal-2018-multimodal, Busso2008-wf, zadeh2016}. Such variability can significantly affect results, making it difficult to isolate the contribution of individual modalities. While prior work acknowledges sampling strategies in passing, they are rarely standardized or reported in a reproducible manner. As a result, observed performance gains may be influenced by temporal redundancy rather than true multimodal fusion capability.
We introduce ERIT, a controlled benchmark for image–text fusion in emotion recognition, constructed from fixed image–text pairs to eliminate video sampling variability. We evaluate models under image-only, text-only, and fusion settings to measure modality contribution under consistent inputs. \footnote{ERIT metadata are available at \url{https://zenodo.org/records/12803448}, and processing and evaluation code at \url{https://github.com/khleeloo/ERIT_Benchmark_Code}. ElderReact source media are not redistributed; users obtain ElderReact separately under its original access conditions and reconstruct the fixed ERIT benchmark locally using the provided code and metadata.}

Our dataset is derived from older adult video data and is designed to support the evaluation and analysis of multimodal emotion recognition in this demographic. ERIT can be used for studying model robustness in settings where emotional expressions differ from those typically represented in standard training corpora.
\noindent \textbf{Contributions.} This paper makes the following contributions:

\begin{itemize}
    \item \textbf{Controlled evaluation setting for multimodal fusion:} We introduce a standardized evaluation setup based on fixed image–text pairs, removing variability introduced by video frame sampling. This enables consistent comparison of modality contributions in human-centered multimodal perception tasks. We instantiate this setting in the ERIT benchmark.
    \item \textbf{Focus on underrepresented demographic:} We provide a benchmark centered on older adult emotional expressions, addressing a gap in existing multimodal emotion recognition benchmarks and enabling analysis of model robustness across populations. Our results reveal that anger and neutral are near-systematically misclassified (recall $\leq$\,0.05), demonstrating that standard models trained predominantly on younger cohorts fail on this demographic in ways that aggregate metrics alone do not expose.
    \item \textbf{Empirical analysis of fusion behavior:} We provide an empirical analysis showing when fusion helps, when it fails, and how current VLMs default to visual priors under ambiguity and modality conflict. ERIT's construction—pairing identical transcripts with images of differing emotion labels—naturally creates cross-modal agreements and conflicts that test whether models genuinely fuse modalities rather than defaulting to a single input.
\end{itemize}


\section{Methodology}

The ERIT benchmark defines fixed image-text pairs derived from videos of older adults reacting to various stimuli. ERIT is designed primarily as a controlled evaluation testbed for multimodal fusion, while also supporting supervised fine-tuning (Table~\ref{tab:erit-all}). Each sample associates a representative frame with transcribed speech and human-annotated emotion labels inherited from ElderReact. The benchmark covers seven emotions: anger, disgust, fear, happiness, sadness, surprise, and neutral. Unlike posed older-adult databases such as FACES and the lifespan database~\cite{Ebner2010, Minear2004}, ERIT captures naturally elicited expressions.




\subsection{Data Collection and Verification} 

The data for the ERIT benchmark was collected from videos of older people reacting to things associated with younger generations, such as various video games, popular music and entertainment, slang, modern technology, etc. We built upon the ElderReact dataset that was labeled for emotions by Amazon Mechanical Turk respondents and created by \cite{Kaixing2019}. 

Since the original ElderReact dataset does not provide ground-truth transcripts, we obtain the textual modality by transcribing all video clips using Whisper Large V3~\cite{Radford2023}. Transcription reliability is estimated using token-level probabilities, where per-word confidence is computed as the mean softmax probability over sub-word tokens. Whisper Large V3 processes all 1,298 videos, producing 18,824 words (14.2 words/video) with a mean confidence of $0.863 \pm 0.143$; 9.8\% of words fall below the 0.7 reliability threshold~\cite{Naderi2024}. Words with confidence $< 0.7$ are flagged in the ERIT metadata for optional filtering. Transcripts are automatically generated and confidence-filtered but are not exhaustively human-verified.

Splits follow the ElderReact source video partitions: all ERIT samples derived from the same source video, and therefore sharing its transcript, remain within the same split. No source video occurs across train, development, and test sets. Participant-level separation cannot be assessed because participant identifiers are unavailable.
\subsection{Frame Selection and Validation}
\label{sec:frame-selection}

The ElderReact video dataset provides human-annotated emotion labels, with one or more labels per video.
To construct static image--text pairs, we select, for each
human-annotated emotion, a representative frame from the video where that emotion is visibly expressed.
\paragraph{Ensemble-based frame selection.} For a video with $k$ human-annotated emotions ${e_1, \dots, e_k}$, we extract $k$ representative frames using an ensemble of three classifiers: \textbf{DeepFace}~\cite{serengil2021lightface}, \textbf{CLIP}~\cite{radford2021learning}, and a ViT-based FER model. For each frame $f$ and emotion $e_i$, we compute a consensus score as the mean predicted probability across $M{=}3$ models and measure inter-model disagreement via variance. The selected frame maximizes consensus. Using heterogeneous models (CNNs, transformers, contrastive learning) reduces bias from any single architecture and mitigates circularity in frame selection.\paragraph{Frame selection ablation.} To assess sensitivity to the selection strategy, we construct two additional variants: \textbf{Random} (uniform sampling per video--emotion pair, seed 42) and \textbf{Midpoint} (temporal center frame). All variants share identical transcripts and labels; only images differ. The ensemble selects frames but does not generate emotion labels; selected frames were not independently re-annotated at frame level. We evaluate LLaVA-NeXT in zero-shot under image-only and fusion settings (Table~\ref{tab:frame-ablation}).
 The substantial variation across frame-selection strategies confirms that frame choice affects downstream evaluation, motivating ERIT's use of a reproducible frame-selection protocol. 
\begin{table}[t]
\centering
\caption{Frame selection ablation on a subset of 576 samples, LLaVA-NeXT zero-shot image-only.}
\label{tab:frame-ablation}
\begin{tabular}{lcc}
\toprule
\textbf{Strategy} & \textbf{Strict Acc.} & \textbf{Multi-label Acc.} \\
\midrule
Random            & 19.79 & 39.24 \\
Midpoint          & 22.40 & 44.44 \\
DeepFace          & 30.38 & 49.83 \\
Ensemble (ours)   & \textbf{31.60} & \textbf{49.65} \\
\bottomrule
\end{tabular}
\end{table}

A distinctive property of ERIT is its high degree of label ambiguity: 1,783 samples (74.9\%) carry multiple valid emotion annotations from the original crowd-sourced labels, with 2.32 ± 1.08 valid labels per sample on average (Table~\ref{tab:dataset}).
\begin{table}[t]
\centering
\caption{ERIT dataset statistics.}
\label{tab:dataset}
\small
\setlength{\tabcolsep}{4pt}
\begin{tabular}{@{}lr@{\;}r@{}}
\toprule
\textbf{Property} & \multicolumn{2}{r@{}}{\textbf{Value}} \\
\midrule
Samples / Videos / Classes & 2{,}379 / 1{,}298 & / 7 \\
Multi-label / Single-label & 1{,}783\,(74.9\%) / & 596\,(25.1\%) \\
Mean Whisper confidence    & \multicolumn{2}{r@{}}{0.87 $\pm$ 0.13} \\
\midrule
\multicolumn{3}{@{}l}{\textit{Emotion distribution (primary label)}} \\
\quad Hap / Sur / Ang      & 730\,(30.7\%) / 556\,(23.4\%) & / 345\,(14.5\%) \\
\quad Dis / Sad / Fea / Neu & 273\,(11.5\%) / 238\,(10.0\%) & / 148\,(6.2\%) / 89\,(3.7\%) \\
\bottomrule
\end{tabular}
\end{table}

\section{Experiments and Results}

We evaluate whether multimodal fusion improves emotion recognition and analyze how vision–language models integrate multimodal cues under controlled image–text inputs. 
We report MAcc as an ambiguity-aware metric, together with strict accuracy, balanced accuracy, macro-F1, and agreement-stratified results. MAcc should not be interpreted as standalone evidence of fusion superiority. We evaluate GPT-4V, GPT-4o~\cite{openai2023gpt4o,openai2023gpt4v}, LLaVA-NeXT~\cite{Liu2024Improved} and Qwen2-VL~\cite{wang2024qwen2vl} in zero-shot settings and fine-tune LLaVA-NeXT with LoRA under image-only, text-only, and fusion inputs. LLaVA-NeXT LoRA uses lightweight rank-8 adapters on the last 4 layers (${\sim}$426K trainable params).
To contextualize ERIT, we compare fusion performance with previously reported results on standard multimodal benchmarks~\cite{Lian2024}. While models such as GPT-4V achieve strong fusion performance on sentiment and emotion datasets (e.g., up to 80.4\% on CMU-MOSI), ERIT yields substantially lower fusion performance at 42.7\%. This gap reflects the controlled nature of ERIT, which removes temporal redundancy from video-based evaluation and requires direct alignment between visual and textual emotion signals, as well as the added difficulty of older adult emotional expression.
\begin{table*}[t]
\centering
\caption{Multi-label and strict accuracy~(\%) on ERIT test ($n{=}576$), with balanced accuracy and macro-F1 for Fusion.
GPT-4V and GPT-4o results are strict accuracy only (multi-label not available for API models).
McNemar's test: LoRA Fusion vs.\ each 0-shot Fusion $p < 0.001$;
LoRA Fusion vs.\ LoRA Image $p = 0.002$; vs.\ LoRA Text $p < 0.001$.}
\label{tab:erit-all}
\begin{tabular}{@{}llcccccccc@{}}
\toprule
 & & \multicolumn{3}{c}{\textbf{Multi-label Acc (\%)}} & \multicolumn{3}{c}{\textbf{Strict Acc (\%)}} & \multicolumn{2}{c}{\textbf{Fusion only}} \\
\cmidrule(lr){3-5}\cmidrule(lr){6-8}\cmidrule(lr){9-10}
\textbf{Model} & \textbf{Type} & Img & Text & Fus & Img & Text & Fus & bAcc & MacroF1 \\
\midrule
GPT-4V          & 0-shot     & --    & --    & --    & 40.14 & 34.48 & 42.30 & --    & --    \\
GPT-4o          & 0-shot     & --    & --    & --    & 38.26 & 30.16 & 41.18 & --    & --    \\
LLaVA-NeXT      & 0-shot     & 49.65 & 29.58 & 60.73 & 30.28 & 16.78 & 33.22 & 59.77 & 27.67 \\
Qwen2-VL        & 0-shot     & 46.19 & 34.43 & 51.73 & 33.22 & 20.42 & 35.47 & 57.98 & 20.78 \\
\midrule
LLaVA-NeXT LoRA & supervised & 57.99 & 64.06 & \textbf{69.97} & 35.94 & 34.90 & \textbf{42.71} & \textbf{59.82} & \textbf{28.67} \\
\bottomrule
\end{tabular}
\end{table*}


In zero-shot settings, fusion consistently improves over unimodal inputs, although the gains over image-only remain modest and the models are visually dominated. LoRA fine-tuning further improves strict accuracy and MAcc, particularly through better use of the textual modality; however, the limited change in balanced accuracy and macro-F1 shows that these gains remain uneven across emotion classes. These results are consistent with prior findings that VLMs without FER fine-tuning achieve 25--40\% on in-the-wild benchmarks~\cite{zhang2024visualprompting}, with further degradation on underrepresented age groups~\cite{kim2021agebias}.

Text-only accuracy on ERIT (16–34\%) is far below standard text-emotion benchmarks (40–62\%)~\cite{Busso2008-wf}. Only 12\% of transcripts contain explicit emotion keywords (e.g., \textit{cry}, \textit{wonderfully}); on these, text accuracy reaches 42\%, versus 34\% on keyword-free ones. Thus, ERIT tests a realistic fusion challenge: models must weight weak, ambiguous text against stronger visual cues.

\begin{figure}[t]
  \centering
  \includegraphics[width=0.95\columnwidth]{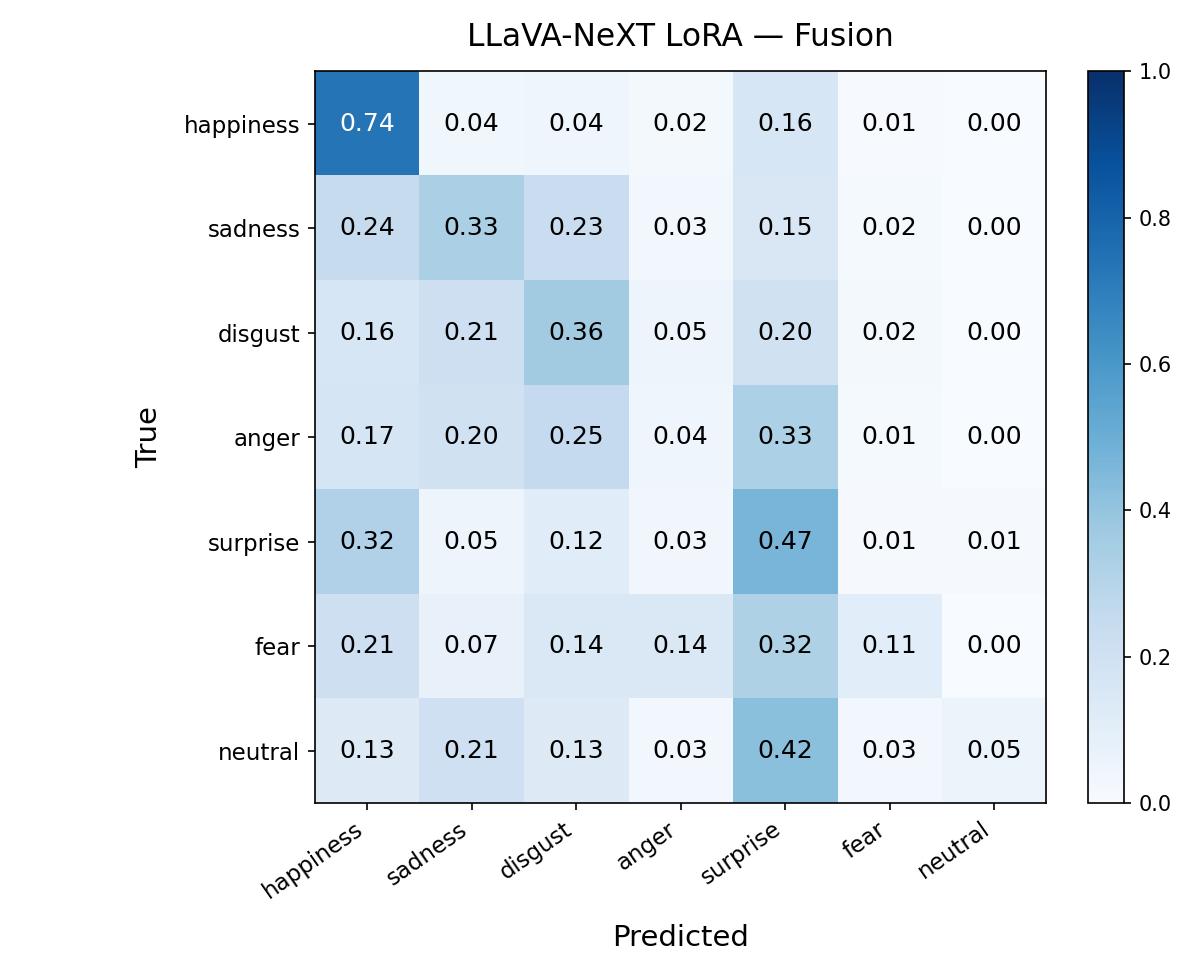}
  \caption{Row-normalized confusion matrix for LLaVA-NeXT LoRA (Fusion of text and image)
    on the ERIT split ($n{=}576$). Each cell shows the fraction of true-class
    samples predicted as the corresponding label (rows sum to~1).}
  \Description{Row-normalized confusion matrix for LLaVA-NeXT LoRA fusion on ERIT subset. Seven emotion classes. Diagonal shows recall: Happiness 0.74, Surprise 0.47, Sadness 0.33, Disgust 0.36, Fear 0.11, Anger 0.04, Neutral 0.05. Off-diagonal: Surprise column is dark, indicating many misclassifications towards Surprise.}
  \label{fig:confusion}
\end{figure}

Figure~\ref{fig:confusion} shows per-class confusion for the fine-tuned model. \textit{Happiness} achieves the highest recall (0.74), while \textit{surprise} acts as a strong attractor, absorbing 33\% of \textit{anger}, 32\% of \textit{fear}, and 42\% of \textit{neutral} samples. Recall for \textit{anger} and \textit{neutral} remains near zero (0.04 and 0.05), consistent with age-related morphological effects (Figure~\ref{fig:aging-morphology}): static features such as drooping eyelids and nasolabial folds mimic negative affect, causing systematic confusion for neutral looking faces.

\subsection{Robustness Analysis}
\label{sec:robustness-results}

We evaluate robustness using five perturbations affecting either modality: no-keyword text, grayscale images, blurred faces, text dropout, and image dropout. We evaluate all non-degenerate combinations across image-only, text-only, and fusion settings, resulting in 14 evaluation cells per model.

\begin{table}[t]
\centering
\caption{Robustness benchmark: LLaVA-NeXT zero-shot across 5
perturbation conditions and 3 modalities.  ``---'' denotes degenerate
cells (perturbation does not affect the active modality).
$\Delta$MAcc is the change from baseline.}
\label{tab:benchmark}
\resizebox{\columnwidth}{!}{%
\begin{tabular}{@{}l rrr rrr@{}}
\toprule
 & \multicolumn{3}{c}{\textbf{Multi-label Acc (\%)}}
 & \multicolumn{3}{c}{\textbf{$\Delta$MAcc (pp)}} \\
\cmidrule(lr){2-4}\cmidrule(lr){5-7}
\textbf{Condition} & I & T & I+T & I & T & I+T \\
\midrule
Baseline       & 59.2 & 37.4 & \textbf{69.5} &   --- &   --- &   --- \\
No-keyword     &  --- & 33.2 & 68.5           &   --- & $-$4.2 & $-$1.0 \\
Grayscale      & 56.6 &  --- & 68.6           & $-$2.6 &   --- & $-$0.9 \\
Blurred faces  & 24.4 &  --- & 55.0           & $-$34.8 & --- & $-$14.5 \\
Text dropout   &  --- &  3.4 & 54.2           &   --- & $-$34.0 & $-$15.3 \\
Image dropout  &  3.4 &  --- & 33.5           & $-$55.8 & --- & $-$36.0 \\
\bottomrule
\end{tabular}}
\end{table}

Three findings emerge.
\textbf{(i)~Fusion is resilient to mild perturbations:}
removing emotion keywords or converting to grayscale degrades fusion MAcc
by only 0.9--1.0\,pp, confirming that fusion draws on complementary cues.
\textbf{(ii)~Facial morphology is critical:}
blurring faces reduces image-only MAcc from 59.2\,\% to 24.4\,\%
($-$34.8\,pp), the largest single-modality drop, yet fusion recovers
to 55.0\,\%, demonstrating that text partially compensates for degraded
visual input.
\textbf{(iii)~Vision dominates:}
image dropout damages fusion far more than text dropout
($-$36.0 vs.\ $-$15.3\,pp), confirming that the visual modality carries
the stronger emotion signal.


\subsection{Label Agreement and Cross-Modal Conflict Analysis}
\label{sec:stratified-results}

ERIT is deliberately constructed so that the same transcribed utterance can be paired with multiple images carrying different emotion labels—some complementary (e.g., \emph{happiness} and \emph{surprise}) and others contradictory (e.g., \emph{sadness} and \emph{neutral})—making text-only recognition unreliable and creating a natural testbed for evaluating whether models genuinely integrate multimodal cues.

\paragraph{Label agreement and modality conflict.}
We split samples by high-agreement (single valid label; $n\!=\!596$, 25.1\%) vs.\ low-agreement (multiple valid labels; $n\!=\!1{,}783$, 74.9\%), and identify conflict samples where image-only and text-only predictions disagree, classified as mild conflicts (related emotions) or strong conflicts (e.g., happiness vs.\ sadness). Table~\ref{tab:stratified} reports performance across these slices.

\begin{table}[t]
\centering
\caption{Stratified analysis (baseline, LLaVA-NeXT 0-shot). MAcc = multi-label accuracy.}
\label{tab:stratified}
\resizebox{0.9\columnwidth}{!}{%
\begin{tabular}{@{}ll rrr r@{}}
\toprule
\textbf{Slice} & \textbf{Subset}
  & \textbf{I} & \textbf{T} & \textbf{I+T} & \textbf{$N$} \\
\midrule
\multicolumn{6}{l}{\textit{Label agreement (MAcc \%)}} \\
  & High (single) & 37.6 & 23.3 & 36.7 & 596 \\
  & Low (multi)   & 66.5 & 42.1 & \textbf{80.5} & 1{,}783 \\
\midrule
\multicolumn{6}{l}{\textit{Modality conflict (MAcc \%)}} \\
& Agree & 54.3 & 54.3 & 65.8 & 622 \\
& Conflict & 61.0 & 31.4 & \textbf{70.9} & 1,757 \\
\bottomrule
\end{tabular}}
\end{table}

Fusion MAcc on low-agreement samples reaches 80.5\%, versus 36.7\% on high-agreement; the larger correct-answer set partly explains this gap, so MAcc should be interpreted as ambiguity-aware rather than as standalone fusion superiority. Across the full ERIT set, 1,757 of 2,379 samples (73.9\%) exhibit image–text prediction conflict, including 980 strong conflicts (55.8\% of conflicts). Fusion improves on conflict samples, with BAcc rising from 21.4 to 29.6 and macro-F1 from 20.2 to 28.6, showing that the effect is not only due to multi-label matching. Mild conflicts reflect related or same-valence disagreements, while strong conflicts denote positive-vs-negative or affective-vs-neutral disagreement. These conflicts are especially important for older-adult affect analysis, where neutral or attentive facial configurations may be misread as active emotion because of age-related facial morphology~\cite{Ebner2009}.


\section{Discussion and Conclusion}
Two central findings emerge from our analysis. First, ERIT's pairing of identical transcripts with images carrying different emotion labels creates natural cross-modal conflicts that test whether models genuinely integrate modalities. Second, current models show systematic errors on older-adult expressions: anger and neutral have near-zero recall, while surprise acts as a strong attractor. These patterns are consistent with facial priors that generalize poorly to older-adult expressions and motivate future work on age-balanced training and class-aware learning objectives. Future extensions of ERIT could incorporate body pose~\cite{Yang2020PosebasedBL} or acoustic prosody~\cite{woolmer2013}.

ERIT provides a controlled testbed for image–text fusion in older adult emotion recognition, pairing fixed visual and textual inputs to enable reproducible analysis of modality contribution. Our experiments show that multimodal fusion improves recognition, while persistent visual dominance and class-specific errors highlight challenges in generalizing to underrepresented demographics. The dataset and protocol support future work on robust, age-aware multimodal perception.

\section{Safe and Responsible Innovation Statement}
ERIT is a research testbed for older-adult emotion recognition, not a deployment system. Key risks include misclassification, privacy concerns, and bias from class imbalance. Human evaluation has not yet been established and remains future work.
ERIT does not redistribute ElderReact source media. Publicly released resources are limited to ERIT-authored code and metadata required to reconstruct the benchmark from an authorized local copy of ElderReact. Users remain subject to the original ElderReact access conditions. ERIT is intended for benchmarking model limitations, not for deployment, surveillance, clinical assessment, or individual profiling. The authors used ChatGPT for language refinement and LaTeX formatting.
\clearpage
\begin{acks}
The work described in this paper was fully supported by a grant from the Research Grants Council of the Hong Kong Special Administrative Region, China (Project Reference No. AoE/E-601/24-N).
\end{acks}

\bibliographystyle{ACM-Reference-Format}
\bibliography{biblio}






\end{document}